\documentclass[acmtog,nonacm]{acmart} 
\usepackage{float}
\usepackage{booktabs}
\usepackage{algorithm}
\usepackage{algpseudocode}
\usepackage{graphicx}
\usepackage{subcaption}
\setcitestyle{nosort,square} %

\usepackage{hyperref}
\usepackage{amsmath}
\usepackage{graphicx}
\usepackage{comment}
\usepackage{kbordermatrix}
\usepackage{multirow,bigdelim}
\usepackage{bm}
\usepackage{array}
\usepackage{wrapfig}
\usepackage{mathtools}
\usepackage{overpic}

\DeclareMathOperator{\FK}{FK_S}

\newcommand{\etal}{et al.}

\newcommand{\bbl}{{\bf L}}
\newcommand{\bbm}{{\bf M}}
\newcommand{\bbo}{{\bf O}}

\newcommand{\bbr}{{\bf R}}

\newcommand{\norm}[1]{\left\Vert #1 \right\Vert_2}

\usepackage{dsfont}
\newcommand{\R}{\mathds{R}} %

\newcolumntype{C}[1]{>{\centering\let\newline\\\arraybackslash\hspace{0pt}}m{#1}}

\newif\ifdraft
\draftfalse 

\ifdraft
\newcommand{\jzc}[1]{{\color[rgb]{0.20,0.60,0.10}[\textbf{Jack:} #1]}}
\newcommand{\rhc}[1]{{\color{blue}[\textbf{Rana:} #1]}}
\newcommand{\kac}[1]{{\color{magenta}[\textbf{Kfir:} #1]}}
\newcommand{\rgl}[1]{{\color{brown}[\textbf{Richard:} #1]}}

\else
\newcommand{\jzc}[1]{}
\newcommand{\rhc}[1]{}
\newcommand{\kac}[1]{}
\newcommand{\rgl}[1]{}

\fi

\newcommand{\ourmethod}{TEDi}
\acmJournal{TOG}

\begin{document}
\makeatletter
\let\@authorsaddresses\@empty
\makeatother
\title{TEDi: Temporally-Entangled Diffusion for Long-Term Motion Synthesis}

\author{Zihan Zhang}
\affiliation{\institution{University of Chicago}}
\author{Richard Liu}
\affiliation{\institution{University of Chicago}}
\author{Kfir Aberman}
\affiliation{\institution{Google Research}}
\author{Rana Hanocka}
\affiliation{\institution{University of Chicago}}

\begin{abstract}
The gradual nature of a diffusion process that synthesizes samples in small increments constitutes a key ingredient of Denoising Diffusion Probabilistic Models (DDPM), which have presented unprecedented quality in image synthesis and been recently explored in the motion domain.
In this work, we propose to adapt the gradual diffusion concept (operating along a diffusion \textit{time-axis}) into the \textit{temporal-axis} of the motion sequence. Our key idea is to extend the DDPM framework to support \textit{temporally varying} denoising, thereby entangling the two axes. Using our special formulation, we iteratively denoise a \textit{motion buffer} that contains a set of increasingly-noised poses, which auto-regressively produces an arbitrarily long stream of frames. With a \textit{stationary} diffusion time-axis, in each diffusion step we increment only the temporal-axis of the motion such that the framework produces a new, clean frame which is removed from the beginning of the  buffer, followed by a newly drawn noise vector that is appended to it. This new mechanism paves the way towards a new framework for long-term motion synthesis with applications to character animation and other domains. Our code and project are publicly available. \footnote{Project page: \url{https://threedle.github.io/TEDi/}}

\end{abstract}

\begin{teaserfigure}
    \centering
    \includegraphics[width=\textwidth]{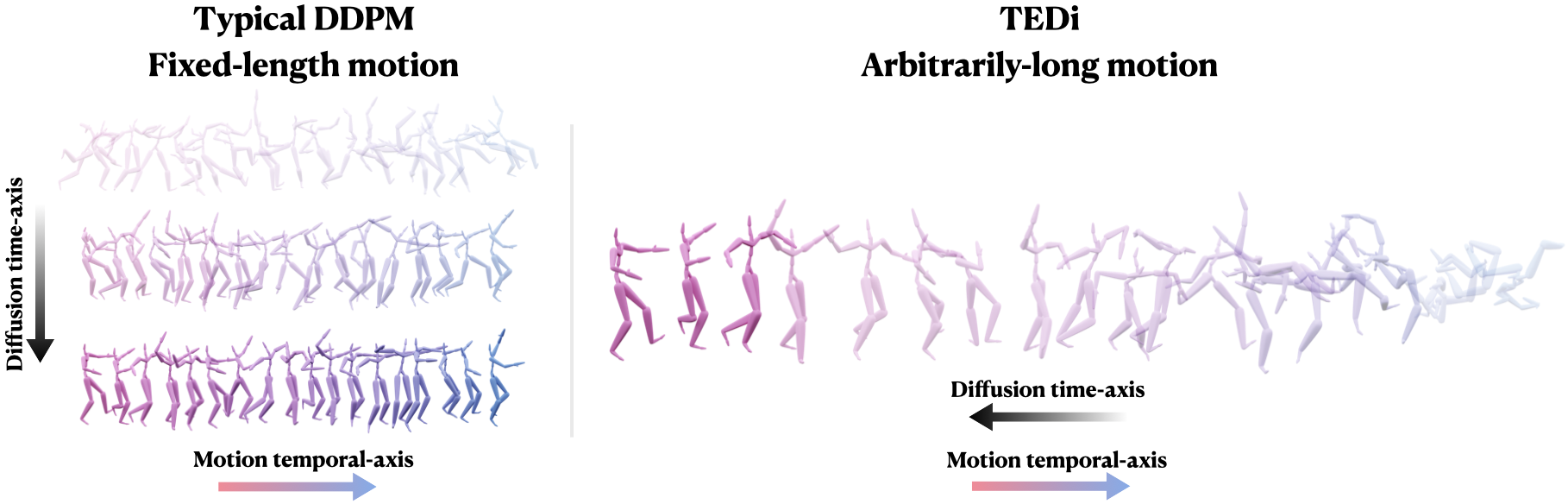} 
    \caption{Inspired by the gradual nature of the diffusion process along a diffusion {\bf time-axis} (left), our approach (right) entangles the {\bf temporal-axis} of motion with the time-axis of the diffusion process (right), enabling a new mechanism for synthesizing arbitrarily long motion sequences.  } 
    \label{fig:teaser}
\end{teaserfigure}

\maketitle

\section{Introduction}

Long-term generation of a motion sequence is a difficult and long standing problem in character animation with myriad applications in computer animation, motion control, human-computer interaction, and more. Generating long-term motion entails producing realistic, non-repetitive sequences which avoids degenerate outputs (i.e., frozen motion).

A promising avenue for generating high-quality motion is through Denoising Diffusion Probabilistic Models (DDPM), which have produced unprecedented quality in image synthesis \cite{ho2020denoising} and have been recently adapted to motion synthesis \cite{tevet2022human, zhang2022motiondiffuse,kim2022flame}. 
A typical adaptation of DDPM to motion synthesis generates a fixed-length motion sequence (i.e., a ``motion image'') from randomly sampled Gaussian noise. 

A fixed-length output is limiting in the context of long-term motion synthesis for a couple of reasons. First, there is no satisfactory approach for creating long-sequences from short-sequences outputs. Simply chaining together motions and blending them may create stitching artifacts. 
Second, a typical diffusion process has limited interactive controllability.  Diffusion requires several hundred denoising iterations before producing a short sequence of clean motions.

We are inspired by the time-dependent nature of the diffusion process, where samples are synthesized from pure noise gradually in small time increments along the diffusion \textit{time-axis}. In this work, we propose to adapt diffusion to the \textit{temporal-axis} of the motion. Our method, referred to as \ourmethod{} (\textbf{T}emporally-\textbf{E}ntangled \textbf{Di}ffusion), extends the DDPM framework by enabling injection of temporally-varying noise levels during each step of the diffusion process, instead of a Gaussian noise with a fixed, temporally-invariant variance.
By \textit{entangling} the \textit{temporal-axis} of the motion sequence with the \textit{time-axis} of the diffusion process, we enable the production of a continuous stream of clean motion frames during each step of the diffusion process.

At the core of our framework lies a \textit{motion buffer}, which encodes noisy future motion frames with varied noise levels. During the training phase, we add temporally varied noise to clean motion sequences, such that each frame has a random level. However, during inference the motion buffer is initialized with a \textit{motion primer} - a sequence of clean motion frames that are being noised with increasing noise levels, such that adjacent frames contains consecutive noise levels.
\ourmethod{} recursively denoises the increasingly-noised future frames. In order to constantly maintain the progressively-noised motion buffer structure during each denoising step, we \textit{insert} a noisy frame at the end of the motion buffer and remove a single clean frame at the beginning. 

This recursive mechanism enables motion sequence frames to be continuously generated, and avoids stitching problems which current motion diffusion models suffer from (see \ref{section:Comparison}).

During inference, we can guide the generation with specific motions by intervening in the process and persistently injecting clean frames, called \textit{guiding motions}. This injection enables us to control and influence the current set of generated frames to \emph{prepare and plan} for the upcoming motion guides. This strategy causes a premeditated and calculated transition between the current frames and the future guiding motions.

Our network continues to denoise an ever-evolving motion buffer, which contains vague information about the future trajectory of the motion sequence. This formulation opens the door to more direct control, and better planning, of the generated motion via manipulation of the motion buffer.
We demonstrate that our framework is capable of producing different types of long motion sequences, and due to its random nature, can provide diverse results even for the same initialization. In addition, we evaluate the model against other long-term generation models. Our experiments show that \ourmethod{} is a natural framework for generating long-term motion sequences.

\section{Related Work}

\subsection{Diffusion Models}
Denoising diffusion probalistic models (DDPMs) and its variants \cite{ho2020denoising,dhariwal2021diffusion,ho2022cascaded} have achieved unprecedented quality on conditional and unconditional image generation, generally surpassing GAN-based ~\cite{dhariwal2021diffusion} methods both in visual quality and sampling diversity. In particular, diffusion models have demonstrated remarkable fidelity and semantic control for text-to-image synthesis and editing tasks when large models are trained on text and image pairs \cite{ramesh2022hierarchical,saharia2022photorealistic,rombach2021highresolution,ruiz2022dreambooth,hertz2022prompt}. In addition, diffusion has been successfully applied in adjacent domains such as text-to-video and image-to-image translation \cite{saharia2022palette}. Moreover, diffusion models are beginning to see increased usage in generative tasks with 3D data. Some recent work enable 3D data generation by reducing it to a 2D task, while others directly train the entire diffusion pipeline on 3D data. 
More recently, in the animation domain,  Zhang~\etal~\cite{zhang2022motiondiffuse}, Kim~\etal~\cite{kim2022flame}, Tevet~\etal~\cite{tevet2022human}, and Shafir~\etal~\cite{shafir2023human} have suggested adapting diffusion models for motion generation by directly applying the diffusion framework, namely by treating the entire motion as an image and denoising all frames in parallel. This adaptation can only generate fixed-length motion sequences which makes long-term generation and interactive control infeasible. In contrast, our framework combines the diffusion framework with an auto-regressive generation scheme, thus enabling generation of arbitrary length sequences by design.

\subsection{Deep Motion Synthesis}

Before the advent of modern deep learning architectures, earlier works attempted to model motion and styles of motion with techniques such as restricted Boltzmann machines \citet{taylor2009factored}. Later on, the seminal set of works by Holden \etal~\shortcite{holden2015learning,holden2016deep} applied convolutional neural networks (CNN) to motion data and learned a motion manifold which can then be used to perform motion editing by, for instance, projection onto the motion manifold. Concurrently, \citet{fragkiadaki2015recurrent} chose to use recurrent neural networks (RNN) for motion modeling. RNN based works also succeeded in short-term motion prediction~\cite{fragkiadaki2015recurrent, pavllo2018quaternet}, interactive motion generation~\cite{lee2018interactive}, and music-driven motion synthesis~\cite{aristidou2021rhythm}. 
\citet{holden2017phase} propose phase-functioned neural networks (PFNN) for locomotion generation and introduce \emph{phase} to neural networks for motion synthesis. Similar ideas are used in quadruped motion generation by \citet{zhang2018mode}. \citet{starke2020local} extended phase to local joints to cope with more complex motion generation. \citet{henter2020moglow} proposed another generative model for motion based on normalizing flow. Additionally, deep neural networks have succeeded in a variety of other motion synthesis tasks such as motion retargeting~\cite{villegas2018neural, aberman2020skeleton, aberman2019learning}, motion style transfer~\cite{aberman2020unpaired, mason2022real}, key-frame based motion generation~\cite{harvey2020robust}, motion matching~\cite{holden2020learned}, animation layering~\cite{starke2021neural} and motion synthesis from a single sequence~\cite{li2022ganimator}.

\subsection{Long-Term Motion Synthesis}

Deep learning models for long term motion synthesis are mostly based on RNNs as they naturally enable auto-regressive generation and capture the time dependencies between animation frames. In general, RNNs have shown much success in natural language processing (NLP) for generating text~\cite{sutskever2011generating}, hand written characters \cite{gregor2015draw}, and even captioning images \cite{vinyals2015show}.
They have also been proposed
for spatio-temporal prediction where \cite{ranzato2014video,srivastava2015unsupervised}
integrated 2D convolutions into the recurrent state transitions of a standard LSTM and proposed the
convolutional LSTM network, which can model the spatial correlations and temporal dynamics in
a unified recurrent unit. Wang~\etal~extended convolutional LSTMs with pairwise memory
cells to capture both long-term dependencies and short-term variations to improve the prediction
quality \cite{wang2017predrnn}. 
 Zhou et al. tackled the problem of error accumulation in long-term random generation by alternating the network's output and ground truth as the input of RNN during training \cite{zhou2018auto}.
This method, called acRNN, is able to generate long and stable motion similar to the training set. However, despite the modified training procedure, acRNN still fails to produce very long motions. One speculation is that acRNN, and RNNs in general, rely on a memory component that is being eroded with time. In contrast, our framework explicitly utilizes frames within our context window which only needs to be the same size temporally as the diffusion time-axis, producing the motion autoregressively in small increments that complies with the successful mechanism of the diffusion process.

\begin{figure}
    \centering
    \includegraphics[width=\columnwidth]{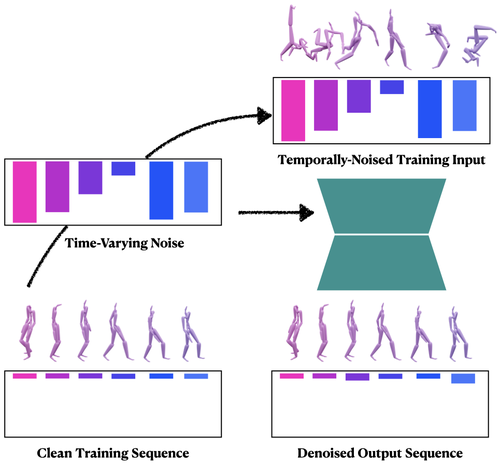}[b]
    \caption{\textbf{\ourmethod{} Training.} We train our diffusion-based model to remove temporally-varying noise that is applied to clean sequences during training. In each iteration we fetch a motion sequence of $K$ frames $[f_1, f_2, \ldots, f_K]$ from the dataset, apply noise to it according to a noise level schedule $[\beta_{t_1}, \beta_{t_2}, \ldots, \beta_{t_K}]$, and train our network to predict the clean motion sequence in a supervised fashion as described in \eqref{eq:diffusion_loss}.}
    \label{fig:overview}
\end{figure}
\section{Method}
We propose a new approach to synthesize long motion sequences using diffusion models. Our approach extends the classic DDPM framework to support injection of temporally-varying noise levels during the diffusion process. This extension enables entangling the \textit{temporal-axis} of the motion sequence with the \textit{time-axis} of the diffusion process. 
In the particular case where the first frame in the sequence is mapped into the lowest noise level, the last frame to the highest level, and the mapping function is linear,  we can continuously synthesize arbitrarily many frames during inference - akin to a \textit{motion buffer}. In each diffusion step we get a clean frame at the beginning of the sequence, shift the frames in the stack by popping the clean frame, and append a new noisy frame (drawn from a Gaussian distribution) to the end of the sequence. Repeating this process during inference results in a new mechanism for long term motion synthesis.
We describe below the motion representation (\ref{sec:motionrep}), novel diffusion framework (\ref{sec:diffusionmodels}), training (\ref{sec:training}) and inference procedure (\ref{sec:inference}). 

\subsection{Motion representation}
\label{sec:motionrep}
We represent a motion sequence by a temporal set of $K$ poses that consists of root joint displacements with respect to the xz-plane ${\bf O_{xz}} \in \R^{K \times 2}$, root joint height $\bf{O_y}\in \R^{K}$, and joint rotations $\bbr \in \R^{K \times JQ}$, where $J$ is the number of joints and $Q$ is the number of rotation features. The rotations are defined in the coordinate frame of their parent in the kinematic chain, and represented by the 6D rotation features ($Q = 6$) proposed by Zhou \etal~\shortcite{zhou2019continuity}. To mitigate foot sliding artifacts, we incorporate foot \emph{contact labels} as a $C\cdot K$ binary values $\bbl\in\{0,1\}^{K\times C}$, which correspond to the contact labels of the foot joints. In our work, we let $C=4$, where the joints are the left(right) heels and toes. All the features are concatenated along the channel axis and we denote the full representation by 
$\bbm \equiv [\bbo_{xz}, \bbo_{y}, \bbr,\bbl]\in \R^{K \times(JQ+C+3)}$. 

\subsection{Diffusion 
Models}
\label{sec:diffusionmodels}

Diffusion Denoising Probabilistic Models (DDPM)~\cite{sohl2015deep,ho2020denoising} are generative models that aim to approximate a given data distribution $q(m_0)$ with an easy and intuitive sampling mechanism that is inspired by diffusion processes in physics. In the particular case of motion synthesis, the data consists of fixed-length motion sequences. During training, the process starts by sampling a clean motion sequence $m_0$ from the dataset, then an IID Gaussian noise is added gradually to form a sequence of noisy motions which constitute the latent variables of the process \{$m_1,\ldots,m_T\}$.
The latent sequence follows $q(m_1,\ldots,m_t\mid m_0)=\prod_{i=1}^{t}q(m_i\mid m_{i-1})$, where 
a sampling step in the forward process (clean data to noise) is defined as a Gaussian transition $q(m_t\mid m_{t-1}):= \mathcal{N}(\sqrt{1-\beta_t}m_{t-1},\beta_t I)$ parameterized by a schedule $\beta_0,\ldots,\beta_T\in (0,1)$. When the total diffusion time step $T$ is large enough, the last noise vector $m_T$ nearly follows an isotropic Gaussian distribution.

In order to sample from the distribution $q(m_0)$, we define the dual ``reverse process'' $p(m_{t-1}\mid m_t)$ from isotropic Gaussian noise $m_T$ to data by sampling the posteriors $q(m_{t-1} \mid m_t)$. Since the intractable reverse process $q(m_{t-1} \mid m_t)$ depends on the unknown data distribution $q(m_0)$, we approximate it with a parameterized Gaussian transition network $p_\theta(m_{t-1}\mid m_t):=\mathcal{N}(m_{t-1}\mid \mu_\theta(m_t,t),\Sigma_\theta(m_t,t))$. 

As suggested by ~\cite{tevet2022human}, instead of predicting the noise as formulated by~\cite{ho2020denoising}, 
we follow~\cite{ramesh2022hierarchical} and the network predicts the signal itself while solving the following optimization problem:

\begin{equation}
    \min_\theta L(\theta):=\min_\theta E_{m_0\sim q(m_0),w\sim N(0,I),t} \norm{m_0  -\mu_\theta(m_t,t)}^2,
    \label{eq:diffusion_loss}
\end{equation}

which maximizes a variational lower bound. In addition, we find that it is best to fix the variance schedule on the reverse process, namely setting $\Sigma_\theta = \beta_t I$ for all time steps, so our model only needs to learn to predict the clean motion. For more details about DDPMs please refer to~\cite{sohl2015deep,ho2020denoising}.

\subsection{Teporally-Entangled Diffusion}
\label{sec:training}

Next, we extend the DDPM framework to support injection of temporally-varying noise levels during the diffusion process. The noise level becomes a function of the frame index and we discard the notion of the diffusion time-axis during training. Effectively, we are setting $T = K$ and identifying the diffusion time-axis and the motion temporal-axis. We propose two schemes for noise injection: 1) random schedule, and 2) monotonic schedule (we avoid the term linear schedule as it is commonly used to indicate a type of variance schedule ~\cite{nichol2021improved}). Note that these are \textit{not} variance schedules. Concretely, given a fixed variance schedule $\beta_{t_i} \in (0, 1), t_i\in \{0, 1, \ldots, T\}$, at each training step the random schedule is given by 
\begin{equation}
    [\beta_{t_1}, \beta_{t_2}, \ldots, \beta_{t_K}],\ t_i \sim \mathcal{U}(0, T).
\end{equation}
On the other hand, the monotonic schedule is given by 
\begin{equation}
    [\beta_{t_1}, \beta_{t_2}, \ldots, \beta_{t_K}],\ t_i =i.
\label{eq:identity}
\end{equation}
The former gives a temporally-varying noise level while the latter gives a monotonically increasing noise level.

In practice, we use a mix of these two noise injection schemes during training, so the model learns to completely denoise a motion sequence with varying noise levels across frames. This enables us to create explicit entanglement between the time axis of the diffusion process and the temporal-axis of the motion - a unique property which will be exploited during inference.

For each iteration during training, we sample from the dataset a motion sequence of length $K$, $[f_1, f_2, \ldots, f_K]$. The model is given the noise injected motion $[\Tilde{f_1}, \Tilde{f_2}, \ldots, \Tilde{f_K}]$ as input where 
$$
    \Tilde{f_i} \sim \mathcal{N}(\sqrt{\bar{\alpha}(t_i)}f_i, (1 - \bar{\alpha}(t_i) I),
$$
for $\bar{\alpha}(t_i) = \prod_{j=1}^{t_j} (1 - \beta_{t_j})$, and is tasked to predict the clean motion $[f_1, f_2, \ldots, f_K]$ directly. To give the network a mixture of the two types of noise injection, we assign $[\beta_{t_j}]_{j=1}^K$ using the random schedule or monotonic schedule with fixed probabilities $p$ and $1-p$. We set $p=\frac{2}{3}$ in practice. In particular, the training objective with the random schedule is similar to those of a pose-oriented diffusion model, where we view the entire motion sequence as a batch of poses with batch size $K$. And at each frame index, the model tries to learn a posterior
$$
    q^*(f^{t-1}\mid f^t)
$$
where the superscript indicates time in the diffusion time-axis, and $q^*(f^0)$ is the data distribution of individual poses. Then, the objective with monotonic noise schedule serves to provide additional supervision to ensure smooth transitions across frames during inference. 

\subsubsection{Loss functions}
As previously mentioned, the benefit of predicting the clean motions directly is that it gives access to regularizations that otherwise would be ill-defined for the mollified distributions. For instance, joint velocities cannot be properly regularized with loss terms for noisy motions. Due to the hierarchical nature of the human model, errors accumulate along the kinematic chains, thus errors on joint rotations should be weighted appropriately with respect to their positions in the hierarchy. Therefore, we add a positional loss loss defined as follows:
\begin{equation}
    \mathcal{L}_{\text{pos}} = \frac{1}{KJ}\sum^K_{t = 1}\norm{\FK(\hat{\bbr}_t, \hat{\bbo}_t) - \FK(\bbr_t, \bbo_t)}^2,
\end{equation}
where $\FK: \R^{JQ}\times \R^3\to \R^{3J}$ is a forward kinematics operator for a fixed skeleton $S$, and $\hat{\bbr}, \hat{\bbo}$ are the model predicted joint rotation and displacements and $\bbr, \bbo$ are the corresponding ground truth. In addition, since foot contact is vital to generating natural motions and enables using inverse kinematics as post-process, we further penalize errors accumulated at the foot joint with the following foot contact loss:
\begin{equation}
    \mathcal{L}_{\text{contact}} = \frac{1}{KC}\sum_{j}\sum_{t=1}^{K-1}\norm{\FK(\bbr_{t+1}, \bbo_{t+1})_j - \FK(\bbr_t, \bbo_t)_j}^2{\cdot} s(\bbl_{tj}),
\end{equation}
where \[
    s = \frac{1}{1 + e^{-12(x-0.5)}}.
\]
This penalizes high foot velocity while having true contact labels, thus ensuring self-consistency of the generated motions. 

\subsubsection{Training}
    In summary, our full training loss is
    \begin{equation}
        \mathcal{L} = \lambda_{\text{diff}}\mathcal{L}_{\text{diff}}+\lambda_{\text{pos}}\mathcal{L}_{\text{pos}}+\lambda_{\text{contact}}\mathcal{L}_{\text{contact}}
    \end{equation}
    where $\mathcal{L}_{\text{diff}}$ corresponds to the diffusion loss specified by equation \eqref{eq:diffusion_loss}, and the $\lambda$ parameters determine the weights of the losses. 
    
Our diffusion network is inspired by the typical U-Net model used in the 2D image diffusion domain~\cite{rombach2022high}. In order for the network to process 1D signals, we use 1D convolutions striding over the temporal axis. We also use 1D attention blocks and skip connections so long term frame correlations are captured within the motion data.

\begin{figure}
    \centering
    \includegraphics[width=\columnwidth]{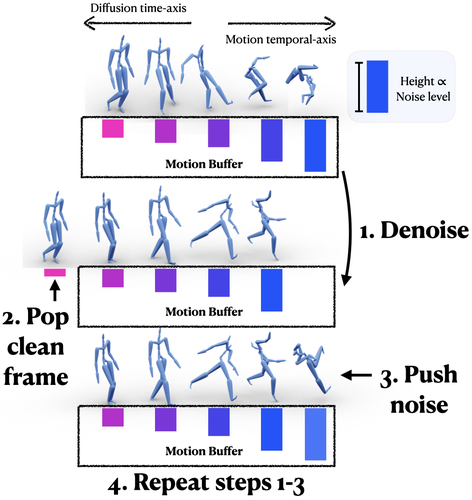}
    \caption{\textbf{\ourmethod{} Recursive Generation.} \ourmethod{} is capable of generating an arbitrarily long motion sequence. First, we initialize our \textit{motion buffer} with a a set of increasingly-noised motion frames. Then (step 1) we denoise the entire motion buffer, (step 2) pop the new, clean frame in the beginning of the motion buffer, and then (step 3) push noise into the end of the motion buffer. This process is repeated recursively. }
    \label{fig:typewriter}
\end{figure}
\subsection{Inference}
\label{sec:inference}

During inference, we take advantage of the monotonic noise schedule that our model trained on. We use a typewriting-like system, as depicted in Fig. \ref{fig:typewriter}. Our model maintains a buffer of frames with monotonically increasing noise, where the first frame in the buffer is mapped to the lowest noise level, and and the last to the highest, as described in~\eqref{eq:identity}. The model is designed to generate motion autoregressively. At the beginning, the buffer is initialized with a given motion sequence that is noised with increasing variance. Then, at each iteration, the model processes all the frames in the motion sequence in parallel and produces a progressively denoised sequence. At this point, the first frame in the sequence is completely clean and can be popped from the buffer. We sample a new frame from standard Gaussian distribution and push it into the motion sequence at the end of the buffer. The model can then iteratively perform this denoising mechanism. This mode of generation can be continued indefinitely as desired, and the resulting motion frames are collected frame by frame from the model output.

    Concretely, let $M_\theta$ be our model and let $I = [\tilde{f}_1, \dots, \tilde{f}_K]$ be the initialization (clean motion that is noised with increasing variance) and let $F_{\text{out}}$ denote the (initially empty) set of output frames. At time step $t$, $t\in \{1, 2, \dots\}$, we have the update
    \begin{align*}
        &F_{\text{out}} = [F_{\text{out}}, M_\theta(I)_1],\\
        &\tilde{f}_{i-1} = M_\theta(I)_{i}, \ i\in \{2, \dots K\}, \\
        &\tilde{f}_K = X \sim \mathcal{N}(0, I),\\
        &I = [\tilde{f}_1, \dots, \tilde{f}_K],
    \end{align*}
    where $M_\theta(I)_i$ denotes the $i$-th frame in the output of our model.

We highlight the distinction from a typical inference pass in the standard diffusion process, which samples Gaussian noise using the full motion length and repeatedly denoises the entire motion. For such a generation scheme, all the frames are required to pass through the model $T$ times. Here, our inference scheme is able to output a new clean frame after only one forward pass of the model. At the same time, a newly sampled frame (pure noise) that gets pushed into the motion buffer will stay in the motion buffer for $T$ iterations, going through all diffusion time steps before getting added to the output. In short, our inference method enables faster autoregressive generation yet ensures that each frame of motion goes through the full diffusion process. 

\section{Experiments}
\begin{figure}
    \centering
    \includegraphics[width=0.5\textwidth]{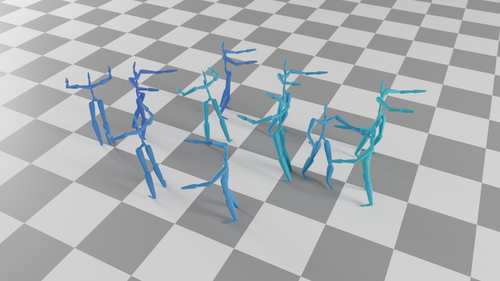}
    \caption{\textbf{Long-term Generation.} Our method synthesizes arbitrarily long motion sequences. In the above figure, we summarize 33 seconds of motion by visualizing the pose every $100$-frames ($\approx$3 seconds). Our model is able to generate plausible motions throughout the entire motion sequence. }
    \label{fig:longterm_generation}
\end{figure}

In this section, we demonstrate the effectiveness of \ourmethod{} on several long-term generation tasks. We show several unique applications of our method, including the ability to plan for upcoming motion using \emph{guided generation}. We also evaluate our method through various comparisons and ablations. For additional qualitative results, please refer to the supplemental video.

\begin{figure*}
    \centering
    \includegraphics[width=0.3\textwidth]{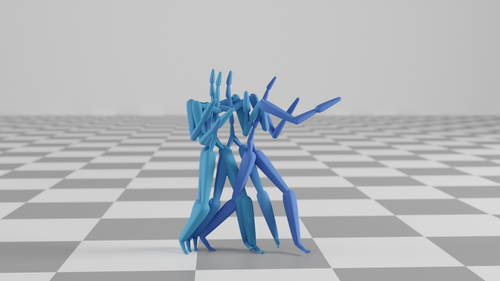}
    \includegraphics[width=0.3\textwidth]{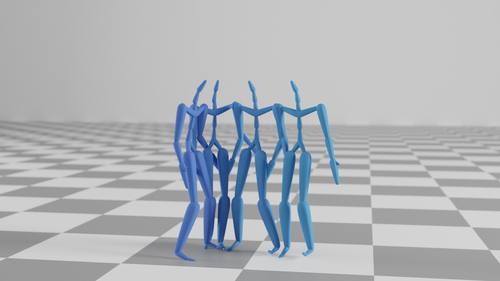}
    \includegraphics[width=0.3\textwidth]{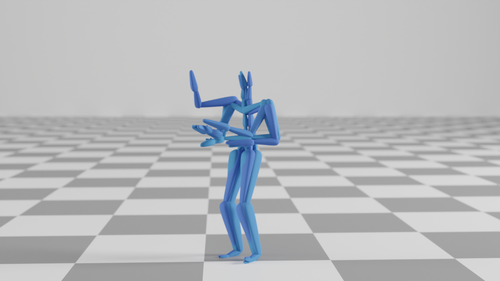}
    \caption{\textbf{Diverse Motions.} Our method is capable of producing a wide variety of long motion sequences. From left to right: Boxing, shuffling, and hand-gestures. }
    \label{fig:uncond_generation}
\end{figure*}

\begin{figure*}
    \centering
    \includegraphics[width=0.3\textwidth]{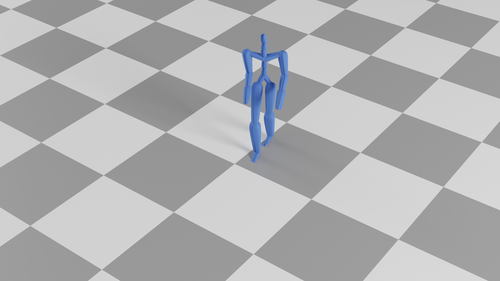}
    \includegraphics[width=0.3\textwidth]{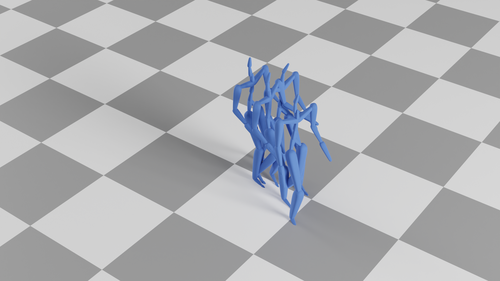}
    \includegraphics[width=0.3\textwidth]{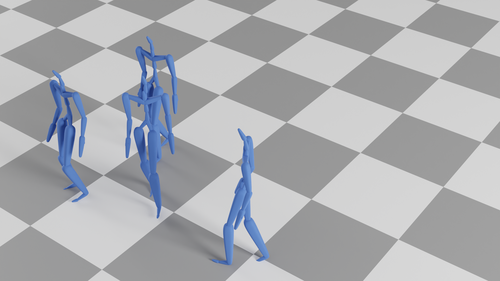}
    \caption{\textbf{Motion Variations.} Due to the stochastic nature of diffusion models, our method is able to generate variations using the same motion primer as input. We show four motions generated from a single primer, from left to right, we can see that the motions begins to differ significantly as time goes on.}
    \label{fig:diversity_generation}
\end{figure*}

\begin{figure*}
    \centering
    \includegraphics[width=0.3\textwidth]{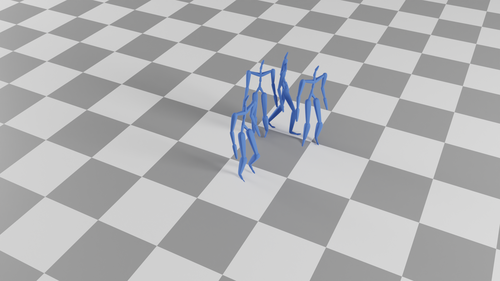}
    \includegraphics[width=0.3\textwidth]{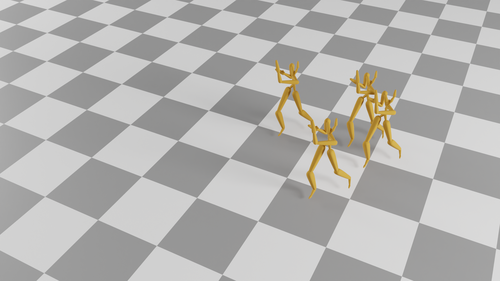}
    \includegraphics[width=0.3\textwidth]{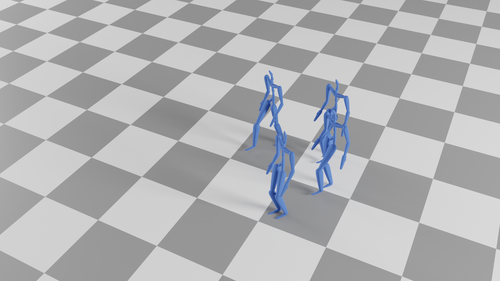}
        \includegraphics[width=0.3\textwidth]{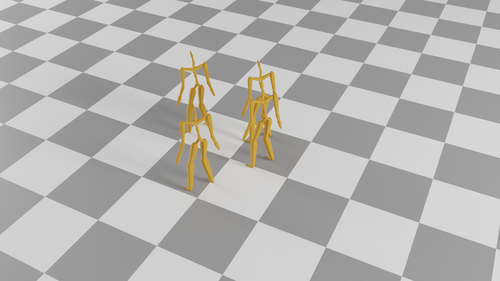}
            \includegraphics[width=0.3\textwidth]{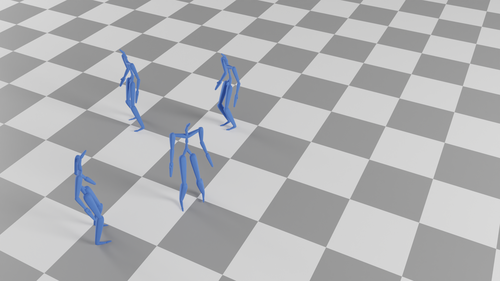}
                \includegraphics[width=0.3\textwidth]{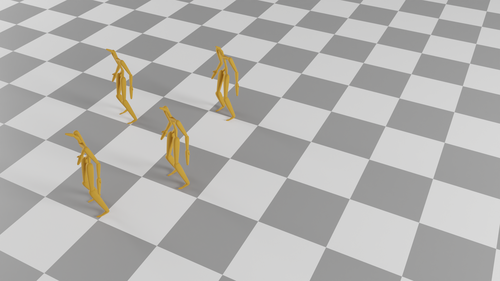}
                
    \caption{\textbf{Guided Generation.} Given a set of \textit{motion guides} $\mathbf{Q}_i$ (shown in yellow), we are able to perform them in sequence at desired points while generating plausible motion in the interactively generated frames (blue). From top-left to bottom-right, our method generates an entire motion sequence that contains the desired motion guides and the interactively synthesized motion. The interactively generated motions will “prepare and plan” for the upcoming motion guides. See the supplementary video. }
    \label{fig:guide_generation}
\end{figure*}

\begin{figure*}
    \centering
    \includegraphics[width=0.3\textwidth]{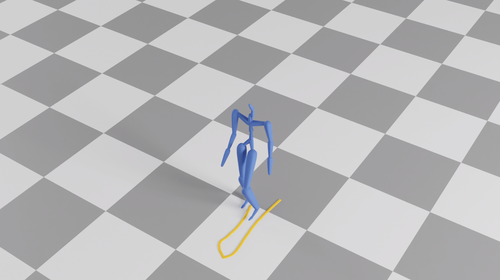}
    \includegraphics[width=0.3\textwidth]{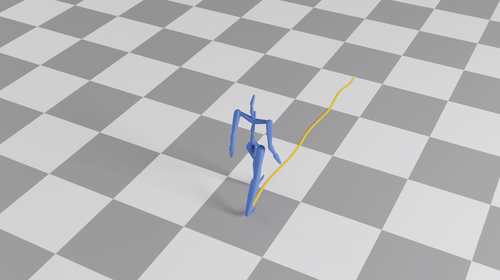}
                
    \caption{\textbf{Trajectory Control.} Similar to guided generation, given the desired trajectory information $\mathbf{P}$ (shown in yellow), our method can generate natural motions that adhere to the given trajectory.}
    \label{fig:traj_control}
\end{figure*}
\subsection{Implementation details}
Our \ourmethod{} framework is implemented with PyTorch, and the training and inference are done on the NVIDIA A40 GPU. We use Adam as our optimizer. For the training data, we use motions from the CMU motion dataset and downsample them from $120$fps to $30$fps. We then sample windows of $500$ frames with a stride of $100$ frames. The CMU dataset contains frames that are shorter than $500$ frames after downsampling, and those are not used for training. Training takes approximately three days for $500$k iterations. 

\subsection{Long-term Generation}
Our \ourmethod{} framework is able to generate long-term motions conditioned on a clean primer motion which is used to populate the initial motion buffer. 
The model is given as input a primer of $K$ frames $\{\Tilde{f_1}, \Tilde{f_2},\ldots, \Tilde{f_K}\}$ which are progressively noised with a monotonic noise schedule. Our iterative inference strategy can then produce an arbitrarily long sequence of new frames. 
We highlight some of the frames from a long-term sequence generated by our method in Fig. \ref{fig:longterm_generation}. The key to maintaining long-term generation is that at each iteration, the newly sampled noise frame ensures that our "buffer" is able to explore new potential motions in the near future, and the iterative denoising process ensures framewise consistency across the motion. In addition, we show in Fig. \ref{fig:uncond_generation} and \ref{fig:diversity_generation} that our method is capable of generating diverse motion sequences.
Full video results can be found in the supplementary video.

\begin{figure}
    \centering
    \includegraphics[scale=0.4]{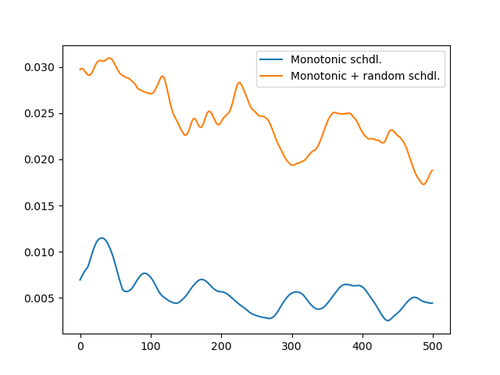}
    \caption{\textbf{Ablations}: Here we show the average motion variance over $500$ frames for our method with and without the random schedule. It can be seen that our random schedule helps avoid motion-collapse. }
    \label{fig:ablation}
\end{figure}

\subsection{Guided generation}

For a character in motion, it is often desired for the character to perform a set of predefined motions which will occur at a point and time in the future. We refer to these frames are \textit{motion guides}. 
Our framework maintains a motion buffer which contains information about the motions to be performed in the future. In order to \emph{influence} the set of currently-generated frames, we directly modify the motion buffer using the motion guide. Specifically, we remove the current set of frames and replace them with a noised version of the motion guide. Then, we discard the predicted denoised frames and replace them with the noised version of the motion guide at the appropriate diffusion time.

Suppose we have a motion buffer of $K$ frames $I = [\tilde{f}_1, \dots, \tilde{f}_K]$ and a set of motion guides $\mathbf{Q}_1, \mathbf{Q}_2, \ldots$ each with length $l_1, l_2, \ldots$ frames that we wish to perform starting at frame number $n_1, n_2, \ldots$, $n_i \geq K$. Assuming we start with the frame number $n=1$ (e.g. the end of the current motion buffer would be frame number $K$), for each predefined motion $\mathbf{Q}_i$, if any of its frames $\mathbf{Q}_{i_j}$, where $ j\in \{1, 2, \ldots, l_i\}$ and $ n_i\leq \mathbf{Q}_{i_j}\leq n_i + l_i$ is such that $n + 5 \leq n_i + j\leq n + K$, then we recursively replace it into the motion buffer. Note that there are five frames right before the start of the motion buffer where we don't recursively replace. This enables the network to smooth out the transitions between the generated frames and the motion guide. We have detailed the procedure for guided generation through recursive replacement in Algorithm \ref{alg:guided_gen}. We demonstrate guided generation in Fig. \ref{fig:guide_generation} and the supplementary material.

\begin{algorithm}
    \caption{Guided generation}
    \label{alg:guided_gen}
    \begin{algorithmic}
        \Require \\
        $M_\theta$: Denoising model\\
        $I = [\tilde{f}_1, \dots, \tilde{f}_K]$: motion buffer \\
        $\{\mathbf{Q}_1, \mathbf{Q}_2, \ldots\}$: motion guides\\
        $\{l_1, l_2, \ldots\}$: motion guide lengths\\
        $\{n_1, n_2, \ldots\}$: starting frame numbers for guidance\\
        $F_{\text{out}} = \varnothing$: ouput frames
        \For{$n$ in $1, 2,\ldots$}
            \State\emph{Evaluate} $M_\theta(I)$
            \For {all frames $Q_{i_j}$}
                \If{$n + 5 \leq n_i + j \leq n + K$}
                    \State $M_\theta(I)_{n_i + j - n} \gets$ $Q_{i_j}$
                \EndIf
            \EndFor
            \State $F_{\text{out}} \gets$ $[F_{\text{out}}, M_\theta(I)_1]$
            \State $\tilde{f}_{i-1} \gets M_\theta(I)_{i}\ \forall i\in \{2, \dots K\}$
            \State $\tilde{f}_K \gets X \sim \mathcal{N}(0, I)$
            \State $I \gets [\tilde{f}_1, \dots, \tilde{f}_K]$
        \EndFor
    \end{algorithmic}
\end{algorithm}

\subsection{Trajectory Control}
Our work can be applied to perform trajectory control during inference without additional training. Similar to the mechanism of guided generation in the previous section, trajectory control also utilizes the inpainting strategy by modifying the motion buffer. Specifically, let $I = [\tilde{f}_1, \ldots, \tilde{f}_K]$ be a motion buffer of $K$ frames, and let $\mathbf{P}\in \R^{3\times N}$ be the trajectory information (root displacements with respect to the xz-plane and root height), where $N$ is the desired number of frames to be generated. During inference, we recursively overwrite the trjactory information in the motion buffer with frames in $P$. The detailed procedure is similar to the one presented in Algorithm \ref{alg:guided_gen}. We demonstrate trajectory control generation in Fig. \ref{fig:traj_control}.

\subsection{Comparison and Ablation}
We next evaluate our approach against alternative baselines, and assess our framework through an ablation study. We refer the reader to the supplementary video attached to this work to assess the results qualitatively. For quantitative evaluation, we assess our ability to avoid collapses in the motion sequence by measuring the variance across all generated frames. In order to measure how non-stationary generated motions are, and to detect the time-point where they collapse, we measure the average variance of poses in a local window.

\subsubsection{Comparison}
\label{section:Comparison}
\begin{figure}
    \centering
    \includegraphics[width=\columnwidth]{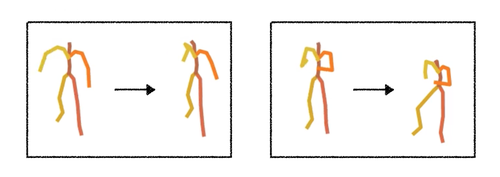}
    MDM
    \vspace{2pt}
     \includegraphics[width=\columnwidth]{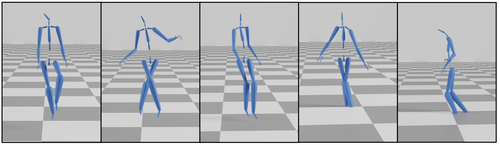}
    ACRNN
    \caption{\textbf{Long-term motion synthesis baseline comparisons.} Top: We show two pairs of \textit{consecutive} frames generated through an in-painting implementation with MDM~\cite{tevet2022human}. Classic in-painting shows visible discontinuity that happens along the border of in-painting. Bottom: ACRNN~\cite{zhou2018auto} when trained on a large dataset is not stable, as seen by the foot levitation and penetration artifacts.}
    \label{fig:quant}
\end{figure}

In this experiment, we focus on comparing our framework to other works on the task of long-term generation. We compare our method with ACRNN ~\cite{zhou2018auto} and the Human Motion Diffusion Model (MDM) ~\cite{tevet2022human}.
In particular, the ACRNN work~\cite{zhou2018auto} is an RNN-based work that receives part of the model's output frames during training, to imitate the inference setting and mitigate motion collapse. MDM is an adaptation of the classic DDPM network for motion generation. While ACRNN is designed to be trained on a subset of samples from the CMU dataset and has long-term generation as default for inference, MDM does not have a default implementation for long term generation. Thus we use a pretrained checkpoint for MDM and implement an inpainting-based scheme to enable long-term generation for MDM. This implementation is the same as the popular "outpainting" technique used in 2D image generation, where we take the latter part of the generated motion and in-paint it to the first part of the generated motion on the next iteration. As in Fig. \ref{fig:quant}, it can be seen that ACRNN is not able to perform well on a large and diverse dataset, producing motions that quickly collapse after initialization. In contrast, \ourmethod{} can produce infinitely long sequences that is robust to collapses. On the other hand, MDM produces significant stitching artifacts along the in-painting boundary. Please refer to the supplemental video for more details.

\subsection{Perceptual Study} We conduct a perceptual study to evaluate the perceived diversity and quality of the generated motions. In addition to MDM and ACRNN, we also add Motion VAE ~\cite{Ling_2020}, a recent autoregressive motion generation model with VAE, as a baseline comparison. Following the setup of DALLE-2 ~\cite{ramesh2022hierarchical}, we show users 3x3 grids of randomly sampled motions from our model, MDM, Motion VAE, and ACRNN, and ask them to choose 1) the set with the most diverse motions and 2) the set with the highest quality motions (only from ours, MDM, or ACRNN). Visual examples of the generated motions in the perceptual study is provided in Appendix \ref{sec:study}

We had 55 respondents for our study, and we report the results in Tab. ~\ref{tab:survey_results}. We conclude from our perceptual study that our method produces motion of equivalent or better quality compared to MDM while significantly outperforming in terms of diversity. 

\begin{table}[htbp]
    \centering
    \begin{tabular}{l|cccc}
    \hline
    & Ours & MDM & ACRNN & MVAE \\
    \hline
    Diversity & \textbf{34} & 12 & 8 & 1 \\
    Quality & \textbf{33} & 17 & 5 & N/A \\
    \hline
    \end{tabular}
    \vspace{5mm}
    \caption{Perceptual study results for our method and baselines.}
    \label{tab:survey_results}
\end{table}

\subsubsection{Ablation}
In Fig. ~\ref{fig:ablation}, we demonstrate the advantage of our training scheme, by training a version of our model with temporally-invariant noise levels. Without temporally varying noise, the network diminishes in both diversity and stability of long-range motion generation.

\section{Conclusion}

In this paper, we proposed \ourmethod{}, an adaptation of diffusion models for motion synthesis which entangles the motion temporal-axis with the diffusion time-axis. This mechanism enables synthesizing arbitrarily long motion sequences in an autoregressive manner using a U-Net architecture. A unique aspect of our work is the notion of a \textit{stationary} motion buffer. Our framework continues to produce clean frames (i.e., progressing along the diffusion-time axis), without \textit{actually} incrementing the diffusion time.
The ability of our pipeline to continually generate motion along the diffusion axis is what enables our framework to robustly and continuously produce novel frames. Interestingly, the ability to naturally use diffusion in such an autoregressive fashion may have implications for other types of sequential data beyond motion, such as audio and video, or modalities where a sequential order can be defined, such as a patch-by-patch order for images.

Our system enables partially-clean-frame to be immediately (or near immediately) popped-off the motion buffer stack. However, a current limitation of our system is that computing a clean from from pure noise requires going through the chain of denoising diffusion.
In the future we are interested in leveraging ideas from DDIM~\cite{song2020denoising} to skip ahead during the denoising process to achieve even lower latency. In addition, our framework may enable future research in long-term text-conditioned motion generation. We are interested in exploring how high-level control may be coupled with low-level user-guidance for the task of long-term generation.

\section*{Acknolwedgements}
We thank the 3DL lab for their invaluable feedback and support. This work was supported in part through Uchicago's AI Cluster resources, services, and staff expertise. This work was also partially supported by the NSF under Grant No. 2241303, and a gift from Google Research.

\bibliographystyle{ACM-Reference-Format}
\bibliography{bibs}

\clearpage
\appendix

\section{Perceptual Study}

\label{sec:study}
\begin{figure}[!h]
    \centering
    \includegraphics[width=\columnwidth]{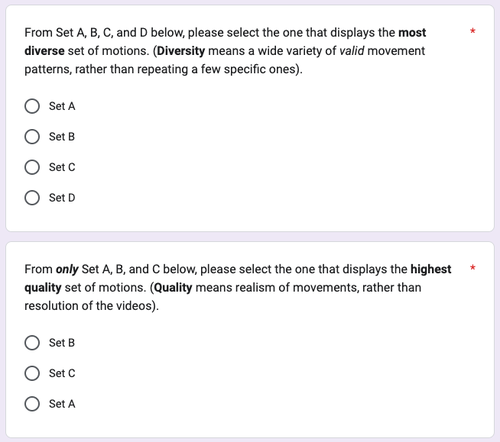}
    \caption{Questions from perceptual study.}
    \label{fig:questions}
\end{figure}

\begin{figure}[!h]
    \centering
    \includegraphics[width=0.7\columnwidth, height=0.23\textheight]{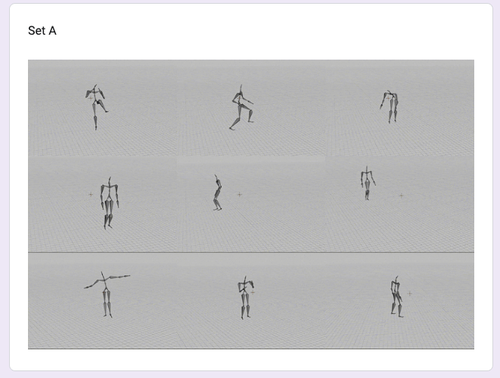}
    \includegraphics[width=0.7\columnwidth, height=0.23\textheight]{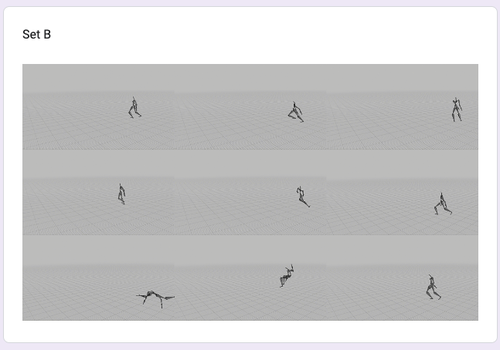}
    \includegraphics[height=0.23\textheight]{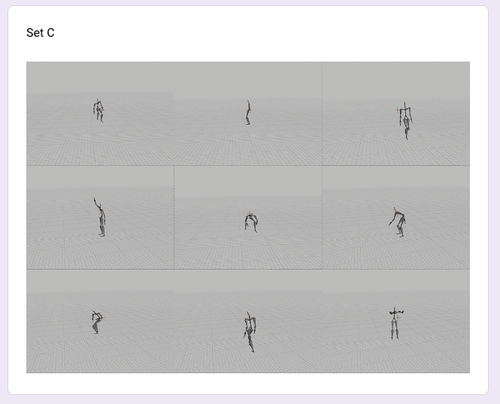}
    \includegraphics[height=0.23\textheight]{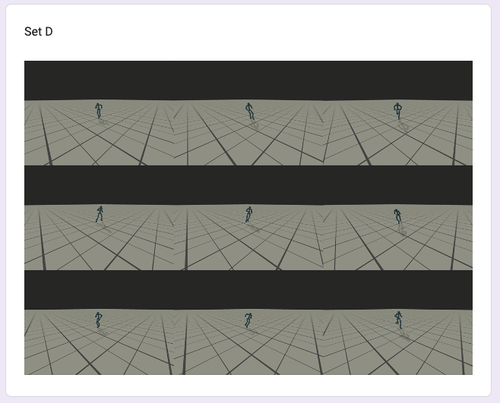}
    \caption{Example motions from perceptual study. \textit{From top to bottom}: Ours, ACRNN, MDM, and Motion VAE.}
    \label{fig:examples}
\end{figure}

Here we provide screenshots of our perceptual study in Fig. ~\ref{fig:questions} and Fig. ~\ref{fig:examples}, as shown for respondents. 

\end{document}